\documentclass[final,5p,times,twocolumn]{elsarticle}
\usepackage[T1]{fontenc}
\usepackage{amsmath,amssymb,booktabs,tabularx,array,graphicx,url}
\usepackage[hidelinks]{hyperref}
\journal{}
\begin{document}
\begin{frontmatter}
\title{Fusion Gains Depend on the Implementation and the Calibration Target: An 18-Task Study of Prediction Intervals with a Single-Source Fallback\tnoteref{status}}
\tnotetext[status]{Author preprint; not peer reviewed.}
\author[a]{Liwei Deng\corref{cor}\fnref{orcid1}}
\ead{liweideng@csuft.edu.cn}
\fntext[orcid1]{ORCID: \url{https://orcid.org/0009-0006-6675-7585}.}
\author[b]{Lu Wang\fnref{orcid2}}
\fntext[orcid2]{ORCID: \url{https://orcid.org/0009-0005-6965-3514}.}
\affiliation[a]{organization={College of Computer and Mathematics, Central South University of Forestry and Technology}, city={Changsha}, postcode={410004}, state={Hunan}, country={China}}
\affiliation[b]{organization={Modern Service College, Hunan Mechanical \& Electrical Polytechnic}, city={Changsha}, postcode={410151}, state={Hunan}, country={China}}
\cortext[cor]{Corresponding author: Liwei Deng.}
\begin{abstract}
A fused model can benefit from an extra learning stage as well as from a second data source. We compare four fusion implementations with matched single-source learners on 18 regression tasks with prediction intervals. Across ten refits under group-wise calibration, linear stacking improved the interval score by 6.0\% against a source model without a second stage and by 0.1\% against its matched single source. The four implementations agreed on 3 of 18 tasks and gave opposite clear verdicts on 12, including 8 of the 10 tasks with naturally distinct sources. With pointwise calibration, they agreed on 5 tasks and gave opposite clear verdicts on 9. A development-set selector with a matched single-source fallback improved ten-refit task-averaged scores by 2.2--6.3\% in every implementation under both calibrations. Adding fused arms to the pool of single-source candidates improved the mean relative score by about 5 percentage points. Under group-wise calibration, the gradient-boosted selector met the criterion frozen before evaluation on eight held-out tasks. In post-opening refits, selectors lost to the matched single source in 4.4--6.1\% of task-level runs under group-wise calibration. Under signed pointwise calibration, three implementations lost in 19--23\% of runs. Fusion evaluations should compare matched candidates on held-out data and report how often the selector loses alongside its average gain.
\end{abstract}
\end{frontmatter}
\section{Introduction}
Information fusion rests on the expectation that a second source adds something the first lacks: a second sensor, a second modality, or a second table of covariates. The evidence offered for a fusion method is usually a comparison in which the fused model scores better than its single-source baselines on held-out data. When the output is a prediction interval rather than a point, the same comparison is made with an interval score.

A single fitted comparison combines the contribution of the second source with the details of training and evaluation. The fused model often has a learning stage that the baseline lacks, so part of its advantage may come from the extra training opportunity rather than from the second source. Its intervals may be calibrated differently from the baseline's. And the result belongs to one implementation and one training sample. Earlier work has raised each concern separately. Multimodal networks can underperform their best unimodal component \cite{r1,r2}, and fusion of noisy, incomplete or imbalanced sources is a recognized difficulty \cite{r3}. Estimated combination weights carry estimation error that can erase the benefit of combining \cite{r4,r5}. Benchmark conclusions based on a single fit can be sensitive to the training sample and to stochastic choices such as parameter initialization \cite{r6}. To interpret a fusion gain as evidence about the sources, we need to know whether it survives a change in the fitted pipeline.

A favorable comparison therefore leaves two questions open. The first is whether the advantage survives a different fusion implementation or a new training sample. The second is how well a rule that is allowed not to fuse performs: a rule that chooses between the fused model and a matched single-source model, and falls back to the single source when fusion does not look better. Choosing between single models and combinations is not a new idea. It has been studied for point forecasts \cite{r7}, for validation-based selection among stacked forecasters \cite{r8}, and for routing individual samples to unimodal experts \cite{r9,r10}. In a targeted search we did not find a study that measures the performance of a matched single-source fallback under calibrated interval scores and paired retraining. Without that measurement, a fusion win and a reliable fusion decision cannot be told apart.

We evaluate two-source prediction intervals on 18 public regression tasks with four fusion implementations that span the usual levels of fusion: late fusion without learning (equal-weight quantile averaging), learned late fusion with a linear and a non-linear combiner (linear and gradient-boosted quantile stacking), and early fusion of the raw features. Ten of the tasks combine naturally distinct sources, such as different sensors or modalities, and eight combine two groups of variables from one table. Each implementation is compared with its own matched single-source learners, which receive the same learning stage with one source withheld. All arms are calibrated by the same rule before scoring: group-wise in the main analysis, and pointwise in a sensitivity analysis. A simple selector chooses between the fused model and the stronger matched single-source model on held-out development data. Eight of the 18 tasks were kept unevaluated until the decision criteria were frozen, and every task was refitted ten times on resampled training data.

The advantage of fusion changed with the baseline, the implementation and the calibration target. Development-set selection with a matched single-source fallback improved the average score in all four implementations under both calibrations. Its loss rate depended on the calibration target.

\begin{itemize}
\setlength{\itemsep}{2pt}\setlength{\topsep}{3pt}
\item Under group-wise calibration, the four implementations reached the same verdict on only 3 of 18 tasks. On 12 tasks, one implementation showed a clear loss from fusion and another a clear gain. Seven of these remained among the three learned implementations alone. The 12 include 8 of the 10 tasks with naturally distinct sources and 4 of the 8 variable-group tasks. Always fusing was clearly worse than the matched single source on 5 to 10 tasks per implementation.

\item On the eight tasks evaluated only after the criteria were frozen, the gradient-boosted stacking selector improved the task-level score on all eight and chose the option with the lower realized test score in all 32 task--state cells.

\item Under group-wise calibration, across four implementations and ten refits per task, the selector scored worse than the matched single source in 4.4--6.1\% of task-level refit runs. Its task-averaged relative improvement was 2.8--6.3\%, and the average over the 18 tasks remained positive in the worst of the ten refit rounds of every implementation.

\item Under pointwise calibration, examined after the sealed tasks were opened, the implementations agreed on 5 tasks and gave opposite clear verdicts on 9. The selector still improved the average score in every implementation, by 2.2--5.8\%, but scored worse than the matched single source in 19--23\% of task-level refit runs for three of the four implementations.

\item A baseline without the learning stage of the fused model misstated the value of the second source. Across refits, linear stacking improved on such a baseline by 5.97\% and on its matched single source by 0.14\%, and verdicts changed in both directions.

\item When the arms of all four implementations were pooled, choosing among the 12 arms instead of the 8 single-source arms improved the average score by about 5 percentage points under both calibration targets.

\end{itemize}
Our contribution is an evaluation design that measures disagreement among fusion implementations separately from the performance of a validation-based fallback rule under retraining.

\section{Related work and positioning}
\textit{Single-source baselines in multimodal learning.} Wang et al. \cite{r1} report that naively fused video networks can score below the stronger unimodal network, and Du et al. \cite{r2} argue that joint training under-learns unimodal features. Krones et al. \cite{r11} relate the gap between unimodal strengths to the gain from fusion in clinical classification. Azizi \cite{r12} studies interval calibration for multimodal regression with missing modalities; there the best single modality is reported by point accuracy only, so the interval results do not compare fusion with a matched single source. Bose et al. \cite{r13} build conformal intervals for multimodal regression from the features of the layer in which the modalities are combined. Comparative studies within fusion research evaluate many fusion methods against one another: Gkoumas et al. \cite{r14} compare eleven fusion strategies for multimodal language analysis on three data sets, and MultiBench \cite{r15} evaluates fusion methods on 15 data sets. Their metrics are point-prediction and classification scores, and the main comparisons of Gkoumas et al. contain no separately trained unimodal baselines. Together these works establish that a single source can match or beat a fused model and that results differ between fusion methods. They do not evaluate a decision rule that may return the single source.

\textit{Selecting between single-source and fused predictors.} Hibon and Evgeniou \cite{r7} compare selecting among individual forecasts and their combinations. Dynamic multimodal fusion routes each sample to a unimodal or a fused expert to save computation \cite{r9}, and Qu et al. \cite{r10} learn a patient-level policy over single-modality and fused predictors. Burenko and Vetrov \cite{r16} allow an ensemble to allocate all members to the stronger modality. These rules operate per sample or per ensemble member, on classification or point prediction, and are judged by accuracy and cost. Here the choice is made at task level between fused and matched single-source interval predictors, scored after common calibration and repeated under paired retraining.

\textit{Forecast combination.} The classical theory gives the optimal linear combination of two forecasts from their error variances and covariance, and implies that an optimally weighted combination is never worse than its better member \cite{r17}. In practice the weights are estimated, and the estimation error can make an estimated combination lose to a simple average \cite{r4,r5}. Blanc and Setzer \cite{r18} use the error ratio, the error correlation and the sample size to decide between the simple average and estimated weights. These studies concern point forecasts, and the choice they analyze is between two ways of combining. We take from them the expectation that a learned combination can fall below its theoretical floor.

\textit{Aggregating probabilistic forecasts.} Schulz et al. \cite{r19} evaluate five ways of aggregating distribution forecasts from deep ensembles on 12 data sets with repeated splits, and Bosch et al. \cite{r8} compare stacking methods for probabilistic time-series forecasting on 50 data sets, including validation-based selection of a single base forecaster. Both show that the method of aggregation matters and that no single method wins everywhere. In both, the members being combined see the same inputs. The reference for a gain is the average member or another forecaster on the same series, not a learner that has been denied one input source. We measure the incremental value of a second input source against a matched single-source learner and evaluate the decision to use it.

\textit{Selection among calibrated prediction sets.} Choosing among prediction sets after they have been calibrated affects their coverage. Yang and Kuchibhotla \cite{r20} show that selecting directly among calibrated candidates gives a coverage bound with a correction for the number of candidates, while recalibrating the selected candidate on data not used for selection restores the nominal marginal guarantee under exchangeability. Liang et al. \cite{r21} develop methods that avoid the additional split. We calibrate each arm before selection and measure the selected intervals' coverage on the test data.

Table 1 summarizes the design elements each study provides.

\begin{table*}[tp]
\caption{Design elements of the closest studies. ``Not established'' means we could not confirm the element from the sections we read.}\label{tab:1}
\scriptsize
\setlength{\tabcolsep}{3pt}\renewcommand{\arraystretch}{1.13}
\begin{tabularx}{\textwidth}{@{}>{\hsize=1.16\hsize\raggedright\arraybackslash}X>{\hsize=1.36\hsize\raggedright\arraybackslash}X>{\hsize=0.82\hsize\raggedright\arraybackslash}X>{\hsize=0.83\hsize\raggedright\arraybackslash}X>{\hsize=0.94\hsize\raggedright\arraybackslash}X>{\hsize=0.89\hsize\raggedright\arraybackslash}X@{}}
\toprule
Study & Second source judged against a matched single-source learner & Interval or probabilistic score & Several fusion implementations & Option not to fuse is evaluated & Paired training resampling with a fixed evaluation set \\
\midrule
Azizi \cite{r12} & Partly: separate source predictors, best single source reported by point accuracy only & Yes & Yes & Single-source point comparison; no fallback rule & No (five seeds) \\
Bosch et al. \cite{r8} & No (different forecasters on the same series) & Yes & Yes & Yes (validation may select one base forecaster; not a single-input-source fallback) & No (rolling validation retraining) \\
Schulz et al. \cite{r19} & No (reference is the mean member score) & Yes & Yes & No & No (repeated splits; no fallback re-evaluated) \\
Xue and Marculescu \cite{r9} & Partly: static unimodal models and unimodal experts & No (accuracy, error and cost) & Yes & Yes (per-sample routing to a unimodal expert) & Not established \\
Qu et al. \cite{r10} & Partly: unimodal baselines & No (AUC and cost) & Yes & Yes (patient-level policy over modality subsets) & Not established \\
Burenko and Vetrov \cite{r16} & Partly: strong unimodal networks or ensembles & No (classification accuracy) & Yes & Yes (allocation may use only the stronger modality) & Not established \\
Gkoumas et al. \cite{r14} & No (main comparisons contain no separately trained unimodal baselines; single-modality ablations for two models) & No (accuracy, F1, error and correlation) & Yes (eleven) & No & No (five training runs per model) \\
This study & Yes & Yes & Yes (four) & Yes (task-level fallback) & Yes (ten paired refits) \\
\bottomrule
\end{tabularx}
\end{table*}
\section{Evaluation design}
Within each implementation, we match the learning opportunity of the fused and single-source arms. Figure 1 shows how we train, calibrate, select and evaluate those arms.

\begin{figure*}[tp]
\centering
\includegraphics[width=0.91\textwidth,height=0.70\textheight,keepaspectratio]{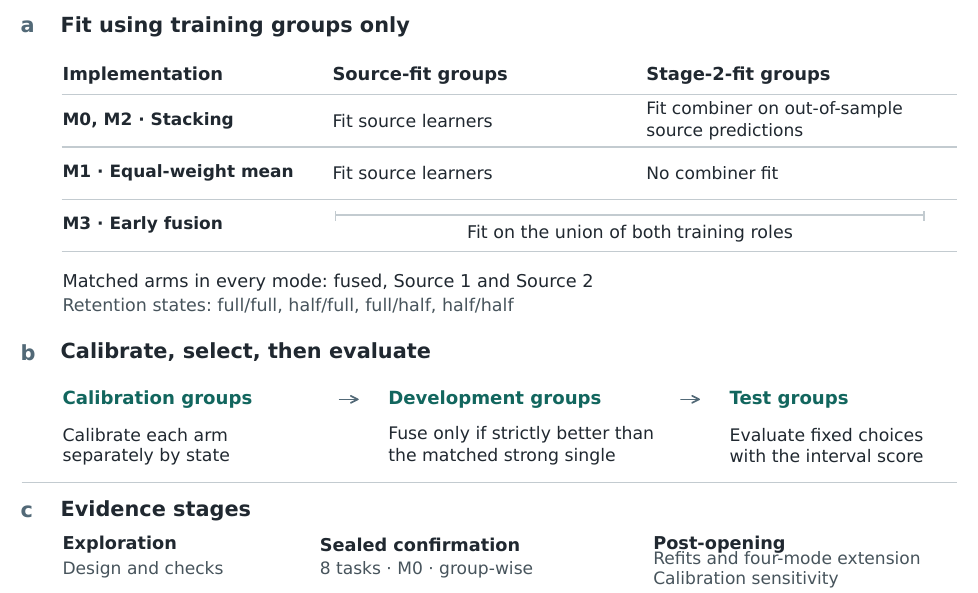}
\caption{Evaluation design and evidence stages. Source-fit and stage-2-fit groups are used only as shown in panel a. The other three group sets are disjoint and used only for calibration, development selection and test evaluation, respectively. Panel c distinguishes sealed confirmation from the post-opening analyses. OpenAI Codex (client versions 0.167.0-alpha.3.2 and 0.169.0-alpha.11.1, OpenAI) assisted with writing the Python plotting code.}\label{fig:1}
\end{figure*}
\subsection{Tasks and sources}
We use 18 public regression tasks from remote sensing, air quality, energy, transport, hydrology, industrial monitoring, real estate, health telemonitoring and multimodal sentiment analysis (Tables 2 and 3). In each task the inputs are divided into two sources, and the division was fixed before any model was fitted. We label a task natural when the two sources are different sensors, modalities or independent observation systems (10 tasks), and variable-group when they are two groups of variables from one table (8 tasks). The labels for the eight sealed tasks (Section 3.7) were assigned from the data audit, without reference to any performance result, and locked before those tasks were evaluated. SRED is described by Azizi and Rudnytskyi \cite{r22}, and the CMU-MOSI features were taken from the MultiBench archive \cite{r15}.

\begin{table*}[tp]
\caption{Tasks and input sources. D denotes naturally distinct sources and V a grouping of variables from one table.}\label{tab:2}
\scriptsize
\setlength{\tabcolsep}{3pt}\renewcommand{\arraystretch}{1.13}
\begin{tabularx}{\textwidth}{@{}>{\hsize=0.77\hsize\raggedright\arraybackslash}X>{\hsize=0.78\hsize\raggedright\arraybackslash}X>{\hsize=0.96\hsize\raggedright\arraybackslash}X>{\hsize=1.59\hsize\raggedright\arraybackslash}X>{\hsize=1.82\hsize\raggedright\arraybackslash}X>{\hsize=0.49\hsize\raggedright\arraybackslash}X>{\hsize=0.59\hsize\raggedright\arraybackslash}X@{}}
\toprule
Task & Data source & Target & Source 1 & Source 2 & Type & Cohort \\
\midrule
BioMassters & \href{https://doi.org/10.57967/hf/1009}{BioMassters} & Mean chip-level above-ground biomass & Monthly Sentinel-1 ASC VV/VH and DSC VV/VH summaries and support indicators & Monthly Sentinel-2 B2--B8, B8A, B11, B12 and CLP summaries and support indicators & D & Exploration \\
Beijing Air Quality & \href{https://doi.org/10.24432/C5RK5G}{UCI Beijing Multi-Site Air Quality} & Station-day mean PM2.5 & Prior-day two-hour summaries of SO\textsubscript{2}, NO\textsubscript{2}, CO and O\textsubscript{3} & Prior-day two-hour summaries of temperature, pressure, dew point, rain and wind & D & Exploration \\
SRED & \href{https://doi.org/10.3390/bdcc6030096}{Swiss Real Estate Dataset} & Log monthly rent & Living space, rooms, latitude, longitude & Pooled text and image embeddings & D & Exploration \\
CMU-MOSI & \href{https://drive.google.com/drive/folders/1uEK737LXB9jAlf9kyqRs6B9N6cDncodq}{MultiBench affect-data archive} & Continuous sentiment score & Text-feature means and standard deviations & Acoustic and visual feature means and standard deviations & D & Exploration \\
Seoul Bike & \href{https://doi.org/10.24432/C5F62R}{UCI Seoul Bike Sharing Demand} & Same-hour rental count & Cyclic hour, weekday and day of year; holiday, operating status and season & Temperature, humidity, wind speed, visibility, dew point, solar radiation, rainfall and snowfall & V & Exploration \\
Four Cities PM2.5 & \href{https://doi.org/10.24432/C52K58}{UCI Five Chinese Cities, Beijing excluded} & PM2.5 (US post) at time t & Prior 24-hour PM mean, SD, min, max, last valid value and valid fraction & Prior 24-hour meteorological means and SDs and wind-direction frequencies & D & Exploration \\
Household Power & \href{https://doi.org/10.24432/C58K54}{UCI Individual Household Electric Power Consumption} & Global active power & Target lags at 60 and 120 min & Local minute of day, weekday and month & V & Exploration \\
Italian Air Quality & \href{https://doi.org/10.24432/C59K5F}{UCI Air Quality} & Reference CO & Four metal-oxide sensor responses & Temperature and relative humidity & D & Exploration \\
Appliances Energy & \href{https://doi.org/10.24432/C5VC8G}{UCI Appliances Energy Prediction} & Appliance energy use & Target lags at 60 and 120 min & Historical outdoor temperature, humidity and wind speed & D & Exploration \\
I-94 Metro Traffic & \href{https://doi.org/10.24432/C5X60B}{UCI Metro Interstate Traffic Volume} & Hourly westbound traffic volume & Local hour, weekday, month and weekend & Temperature, rain, snow and cloud cover at t \ensuremath{-} 2 h & V & Exploration \\
CAMELS-GB & \href{https://doi.org/10.5285/8344e4f3-d2ea-44f5-8afa-86d2987543a9}{CAMELS-GB v1, gauge 41004} & Daily specific discharge & Precipitation at t and t \ensuremath{-} 1 day & Temperature, potential evapotranspiration, radiation, humidity and wind speed & D & Sealed \\
SM2RAIN & \href{https://doi.org/10.5281/zenodo.3766792}{CNR-IRPI SM2RAIN collocations} & Gauge-based daily rainfall & ASCAT and SMAP relative soil moisture & ERA5 evapotranspiration at t and t \ensuremath{-} 1 day & D & Sealed \\
Capital Bikeshare & \href{https://doi.org/10.24432/C5W894}{UCI Bike Sharing Dataset} & Hourly total rentals & Local hour, weekday, month and weekend & Weather category, temperature, humidity and wind speed at t \ensuremath{-} 2 h & V & Sealed \\
Birmingham Parking & \href{https://doi.org/10.24432/C51K5Z}{UCI Parking Birmingham} & Occupancy of designated car parks & Lagged occupancy of two peer car parks & Scheduled local hour, weekday, month and weekend & V & Sealed \\
Tetouan Power & \href{https://doi.org/10.24432/C5B034}{UCI Power Consumption of Tetouan City} & Zone 1 power consumption & Target lags at 60 and 120 min & Historical temperature, humidity and wind speed & D & Sealed \\
Steel Energy & \href{https://doi.org/10.24432/C52G8C}{UCI Steel Industry Energy Consumption} & Energy use in kWh & Target lags at 60 and 120 min & Local minute of day, weekday and month & V & Sealed \\
Gas Turbine NOx & \href{https://doi.org/10.24432/C5WC95}{UCI Gas Turbine CO and NOx Emission Data} & Same-hour NOx concentration & Ambient temperature, pressure and humidity & Six turbine operating variables & V & Sealed \\
Parkinson Telemonitoring & \href{https://doi.org/10.24432/C5ZS3N}{UCI Parkinsons Telemonitoring} & Interpolated motor UPDRS & Voice measures (jitter, shimmer, NHR, HNR, RPDE, DFA, PPE) & Age, sex and time since recruitment & V & Sealed \\
\bottomrule
\end{tabularx}
\end{table*}
\begin{table*}[tp]
\caption{Evaluation roles by task. Time ranges are the extremes of the retained records.}\label{tab:3}
\scriptsize
\setlength{\tabcolsep}{3pt}\renewcommand{\arraystretch}{1.13}
\begin{tabularx}{\textwidth}{@{}>{\hsize=0.52\hsize\raggedright\arraybackslash}X>{\hsize=1.27\hsize\raggedright\arraybackslash}X>{\hsize=1.21\hsize\raggedright\arraybackslash}X@{}}
\toprule
Task & Rows/groups: source-fit; stage-2-fit; development; calibration; test & Group unit; time range \\
\midrule
BioMassters & 3,893/3,893; 2,118/2,118; 874/874; 889/889; 909/909 & Spatial leakage component; no common calendar \\
Beijing Air Quality & 4,146/392; 5,537/483; 1,768/151; 2,111/180; 2,770/240 & Target date across stations; 2013-03-02 to 2017-02-28 \\
SRED & 3,998/3,998; 2,998/2,998; 1,499/1,499; 1,501/1,501; 1,109/1,109 & Listing; no listing-level dates \\
CMU-MOSI & 634/26; 511/21; 138/5; 214/9; 686/31 & Video; video-relative indices only \\
Seoul Bike & 4,368/182; 1,728/72; 864/36; 840/35; 864/36 & Date; 2017-12-01 to 2018-11-30 \\
Four Cities PM2.5 & 19,947/401; 36,266/437; 19,473/218; 20,210/218; 19,645/219 & Date across cities; 2011-11-23 to 2015-12-31 \\
Household Power & 129,465/90; 47,877/35; 27,240/19; 25,680/18; 25,791/18 & Date; 2007-01-01 to 2007-06-30 \\
Italian Air Quality & 3,522/173; 1,509/69; 760/35; 745/35; 808/35 & Date; 2004-03-10 to 2005-04-04 \\
Appliances Energy & 9,822/69; 3,870/27; 1,998/14; 1,998/14; 1,963/14 & Date; 2016-01-11 to 2016-05-27 \\
I-94 Metro Traffic & 15,826/920; 7,580/368; 4,017/184; 4,010/184; 4,059/185 & Date; 2012-10-02 to 2018-09-30 \\
CAMELS-GB & 6,993/6,993; 2,796/2,796; 1,398/1,398; 1,398/1,398; 1,398/1,398 & Date; 1970-10-02 to 2015-09-30 \\
SM2RAIN & 460,706/501; 172,113/200; 93,223/99; 94,773/99; 91,527/100 & UTC date across grid points; 2015-04-03 to 2017-12-30 \\
Capital Bikeshare & 7,855/365; 3,192/146; 1,606/73; 1,606/73; 1,585/73 & Date; 2011-01-01 to 2012-12-31 \\
Birmingham Parking & 2,977/35; 1,205/14; 377/5; 765/9; 693/8 & Date across car parks; 2016-10-04 to 2016-12-19 \\
Tetouan Power & 26,196/182; 10,356/72; 5,316/37; 5,172/36; 5,316/37 & Date; 2017-01-01 to 2017-12-30 \\
Steel Energy & 17,464/182; 7,000/73; 3,544/37; 3,448/36; 3,544/37 & Date; 2018-01-01 to 2018-12-31 \\
Gas Turbine NOx & 18,231/110; 7,320/44; 3,689/22; 3,630/22; 3,856/23 & 168-record blocks within year; 2011--2015, no row-level dates \\
Parkinson Telemonitoring & 2,892/21; 1,127/8; 279/2; 1,280/9; 297/2 & Patient; patient-relative days, no common calendar \\
\bottomrule
\end{tabularx}
\end{table*}
\subsection{Data roles}
Each task is split by group into five disjoint roles: source-fit, stage-2-fit, development, calibration and test. Each predefined group is kept intact, so no group contributes to more than one role. Grouping units are task-specific and include dates, spatial components, listings, videos, within-year record blocks and patients (Tables 2 and 3). For the 12 newly qualified tasks the nominal allocation of groups was 50\%, 20\%, 10\%, 10\% and 10\%, in chronological or hash order as appropriate, with boundary purges and minimum-group adjustments where required. Seoul Bike and Four Cities PM2.5 use the same nominal proportions over their calendar spans. BioMassters, Beijing, SRED and CMU-MOSI retain task-specific spatial, calendar or inherited-fold splits.

For M0 and M2, the source learners are trained on the source-fit role, and their out-of-sample predictions on the stage-2-fit role supply the stacking inputs. M1 fits no combiner, and M3 is trained directly on the union of the two fitting roles (Section 3.4). In every implementation the development role is used only to choose between arms, the calibration role only to estimate interval expansions, and the test role only to score the fixed choices. None of the three is used to fit a learner.

For 13 tasks all roles lie on a common calendar and the development data precede the test data. For the gas-turbine task the roles are separated by year, with development in 2014 and test in 2015. For four tasks (BioMassters, SRED, CMU-MOSI and Parkinson Telemonitoring) the records carry no common calendar, and the order of development and test data cannot be established.

\subsection{Information-retention states}
Real deployments do not always receive both sources complete. Each arm is therefore evaluated in four states, in which each source is either fully or half retained: FULL--FULL, HALF--FULL, FULL--HALF and HALF--HALF, the first entry referring to source 1. For a half-retained source we keep the predefined feature groups at alternating positions (the first, third, fifth and so on) and set every coordinate in the intervening groups to missing. The mask is deterministic. It retains alternating feature groups, which is not necessarily half of the columns or half of the information. The second-stage fits of M0 and M2 and the direct fit of M3 see the four states with equal weight. A task therefore contributes four task--state cells, 72 cells in all.

\subsection{Four fusion implementations and their matched single-source learners}
The four implementations span fusion without learning, learned late fusion with a linear and a non-linear combiner, and early fusion (Table 4). In every implementation the single-source arms receive the same learning stage as the fused arm, with one source withheld. The matching is within an implementation; the four implementations do not share a training budget. All implementations take fixed, precomputed features for each source, and no representation is learned jointly across sources. Differences between implementations therefore do not involve joint representation learning, whose interaction with fusion is documented for multimodal networks \cite{r1,r2}.

\begin{table*}[tp]
\caption{The four fusion implementations and their matched single-source arms.}\label{tab:4}
\footnotesize
\setlength{\tabcolsep}{3pt}\renewcommand{\arraystretch}{1.13}
\begin{tabularx}{\textwidth}{@{}>{\hsize=0.61\hsize\raggedright\arraybackslash}X>{\hsize=1.15\hsize\raggedright\arraybackslash}X>{\hsize=1.24\hsize\raggedright\arraybackslash}X@{}}
\toprule
Implementation & Fused arm & Matched single-source arms \\
\midrule
M0 Gradient-boosted quantile stacking & A gradient-boosted quantile model fitted on the quantile summaries of both source models & The same model fitted on the quantile summaries of one source model \\
M1 Equal-weight quantile averaging & The level-wise mean of the two source models' quantiles & Each source model's own quantiles \\
M2 Linear quantile stacking & A linear quantile regression on the summaries of both source models & The same regression on the summaries of one source model, fitted on the same rows \\
M3 Early fusion & A gradient-boosted quantile model fitted on the concatenated raw features & The same model fitted on the raw features of one source \\
\bottomrule
\end{tabularx}
\end{table*}
All models predict the 0.05, 0.50 and 0.95 quantiles. Where predicted quantiles cross, they are made monotone by row-wise isotonic projection. M0 uses histogram-based gradient boosting for the source quantiles and for the stacking stage. M1 averages the corresponding quantiles of the same two source predictors and fits no second stage. M2 uses unregularized linear quantile regression with an intercept on the same summaries as M0, with a common cap of 12,288 stage-2 rows before state expansion; when the cap applies, whole groups are retained. M3 fits histogram-based gradient boosting to the masked raw features on the union of the source-fit and stage-2-fit roles. All gradient-boosting fits use at most 150 iterations, at most 31 leaves, at least 20 samples per leaf, a learning rate of 0.1, no L2 regularization and no early stopping. The software environment was Python 3.14.4, NumPy 2.5.1, SciPy 1.18.0 and scikit-learn 1.9.0.

\subsection{Calibration and scoring}
Raw quantile intervals from different arms are not comparably calibrated, and an uncalibrated comparison confounds the value of the second source with calibration differences between arms (Section 4.1). Every arm is therefore calibrated separately in each state before scoring. Let $l$ and $u$ be the predicted 0.05 and 0.95 quantiles of an observation $y$. For each group $g$ of the calibration role we compute the largest distance by which any of its observations falls outside its interval:

\begin{equation}
r_g = \max_{i \in g} \max\{0,\; l_i - y_i,\; y_i - u_i\}\label{eq:1}
\end{equation}
With $G$ calibration groups, the expansion $c$ is the \ensuremath{\lceil}0.90($G$ + 1)\ensuremath{\rceil}-th smallest group residual. The calibrated interval is [$l$ \ensuremath{-} $c$, $u$ + $c$], and the median is unchanged. The nominal level is 90\%. The same expansion is applied to the development and test predictions of that arm and state.

This rule targets group-wise coverage: if calibration and test groups are exchangeable, all observations of a new group fall inside their calibrated intervals with probability at least 90\%, so pointwise coverage is typically higher than 90\%. It can also only widen intervals, so an arm whose raw intervals are too wide is not corrected. A common calibration rule therefore does not give the arms identical realized coverage. The selector chooses among arms that are already calibrated, using held-out development scores, and the chosen arm is not recalibrated. The nominal guarantee for a prespecified arm therefore need not carry over to the selected intervals \cite{r20,r21}, and we report their empirical test coverage.

Realized coverage reflects the group-wise target. After calibration, the median task-level pointwise coverage lay between 99.5\% and 99.9\% for every implementation and strategy, and task averages ranged from 83.8\% to 100\% (Supplementary Tables S6.3 and S6.5). The calibrated intervals are therefore much wider than pointwise 90\% intervals, and the calibrated interval score is driven mainly by width. Relative to the matched strong single source, fusion showed no consistent direction of change in pointwise coverage, group-wise coverage or width; the median ratio of fused to single-source width ranged from 0.977 to 1.024 across implementations. No arm or state lacked a valid calibration rank.

Because the group-wise target is conservative for individual observations, Section 4.6 repeats the main outcome readouts with pointwise calibration.

Intervals are scored with the interval score at nominal level 90\% \cite{r23}, which adds the interval width to a penalty proportional to the distance by which an observation falls outside the interval. Lower is better. For a lower bound $l$, an upper bound $u$, an observation $y$ and $\alpha$ = 0.10:

\begin{equation}
S(l,u;y) = (u-l) + \frac{2}{\alpha}(l-y)\,\mathbf{1}\{y<l\} + \frac{2}{\alpha}(y-u)\,\mathbf{1}\{y>u\}\label{eq:2}
\end{equation}
The score of an arm in a state is the mean of this quantity over the test observations. Targets differ in unit across tasks, so strategies are compared by a relative score difference against the stronger matched single source of the same implementation:

\begin{equation}
d = \frac{S_{\text{strategy}} - S_{\text{strong single}}}{S_{\text{strong single}}}\label{eq:3}
\end{equation}
Negative $d$ means the strategy scores better than the single source. The task-level $d$ is the mean over the four states, and averages across tasks weight tasks equally. For a single fit we attach a 95\% percentile band to each task-level $d$ from 1,000 paired resamples of the test groups, holding the fitted models, the calibration thresholds and the development-set choices fixed. We call a loss or gain clear when this band lies wholly above or wholly below zero. The band describes test-sample variation for one fit. Section 3.8 measures variation across training samples by refitting.

\subsection{Strategies}
Three strategies are compared within each implementation and state.

\begin{enumerate}
\item \textit{Always strong single.} The single-source arm with the lower calibrated score on the development role is used. This is the reference, with $d$ = 0.

\item \textit{Always fuse.} The fused arm is used.

\item \textit{Selector.} The fused arm is used if its calibrated development score is strictly lower than that of the strong single source; otherwise the strong single source is used. The choice is made once on the development role and is not revisited on the test role.

\end{enumerate}
\subsection{Status of the evidence}
The study separates exploration, sealed confirmation and analyses conducted after opening the sealed tasks.

\textit{Exploration (10 tasks).} Six tasks had been studied in earlier phases of this project. Twelve further tasks passed a data-eligibility audit conducted without predictive performance evaluation, and four of them were randomly assigned to the exploration set. Analyses on these ten tasks were unrestricted and are logged in full.

\textit{Sealed confirmation (8 tasks, M0).} The remaining eight qualified tasks were set aside. No performance was computed on them in this project before two criteria were frozen on 11 November 2026: one on the selector (Criterion 1) and one on always fusing (Criterion 2), given in Table 5. The tasks were then run once with gradient-boosted stacking. Only these two criteria are confirmatory.

\textit{Post-opening extensions (all 18 tasks).} After the sealed tasks had been opened, we added ten refits per task for M0 and then the three further implementations with the same refits. Each extension had its readouts and decision rules frozen before it ran, but both re-use tasks whose results were already known, and we report them as robustness evidence. A third post-opening analysis, also frozen before it was computed, repeats the main outcome readouts with pointwise calibration (Section 4.6). The distinction between fusion wins and fusion decisions that organizes this paper was formulated after the extensions.

Protocol files, freeze times and hashes for each stage are listed in Supplementary Table S1.1.

\subsection{Refits}
To measure sensitivity to the training sample, every task was refitted ten times for every implementation. In each refit the groups of the source-fit and stage-2-fit roles were resampled with replacement and all models were trained again; the development, calibration and test roles were unchanged. The same ten resamples were used for all four implementations, so results are paired across implementations. With the original fit this gives 11 fits per task and implementation, and 180 refit runs per implementation. For each refit we record whether the sign of the task-level $d$ for always fusing matches the single-fit verdict, and whether the selector scored better than, equal to or worse than the strong single source.

OpenAI Codex (OpenAI; client version 0.167.0-alpha.3.2 in the archived analysis sessions) assisted with writing Python scripts for model fitting and training-group refits, interval calibration and scoring, development-set selection, and result tabulation. For the sealed confirmation and the four-implementation extension, separate verification scripts recomputed selector choices, interval-score differences, and the reported paired test-group bootstrap bands from saved predictions and checked them against the archived results.

Figure 1 is a schematic. Figures 2 to 4 were rendered with Matplotlib 3.11.0 from the archived single-fit verdicts, refit-round means and selector loss counts, respectively. Supplementary Figure S1 uses the archived task-level stability and loss counts. The plotting scripts and derived inputs are included in the reproducibility package. OpenAI Codex (client versions 0.167.0-alpha.3.2 and 0.169.0-alpha.11.1, OpenAI) assisted with writing the Python plotting code.

\section{Results}
\subsection{Sealed confirmation}
Both frozen criteria were met on the eight sealed tasks (Table 5). The selector improved the task-level score on every task: the upper ends of the eight bands ranged from \ensuremath{-}8.09\% to \ensuremath{-}0.11\%, all below zero. In the 32 task--state cells, the fused arm had the higher test score in 9 and the lower in 23. The selector fell back to the single source in all 9 and chose the fused arm in all 23. The relative score differences between the fused arm and the strong single source on development and on test data, each normalized by the single-source score on the same data, agreed in sign in every cell and were closely ranked (Spearman \ensuremath{\rho} = 0.95).

\begin{table*}[tp]
\caption{Frozen criteria and results on the eight sealed tasks (M0).}\label{tab:5}
\footnotesize
\setlength{\tabcolsep}{3pt}\renewcommand{\arraystretch}{1.13}
\begin{tabularx}{\textwidth}{@{}>{\hsize=1\hsize\raggedright\arraybackslash}X>{\hsize=1\hsize\raggedright\arraybackslash}X>{\hsize=1\hsize\raggedright\arraybackslash}X@{}}
\toprule
Frozen criterion & Threshold & Result \\
\midrule
1a. Tasks on which the selector is clearly worse than the strong single source & At most 1 of 8 & 0 of 8 \\
1b. Fallback rate in cells where the fused arm scored worse on test & At least 75\% & 9 of 9 \\
1c. Selection rate in cells where the fused arm scored better on test & At least 60\% & 23 of 23 \\
2. Tasks on which always fusing is clearly worse, and clearly better & At least 2 of each & 2 worse, 5 better \\
\bottomrule
\end{tabularx}
\end{table*}
Always fusing was clearly worse than the matched single source on two tasks, Birmingham Parking (+5.37\%, band 3.97\% to 6.49\%) and Tetouan Power (+3.60\%, band 3.35\% to 3.83\%). It was clearly better on five, and its band straddled zero on one. Averaged over the eight tasks, always fusing improved the score by 3.13\% and the selector by 4.81\%. Always fusing therefore incurred two clear losses, while never fusing would have forgone five clear gains.

Calibration changed the verdict on fusion in many cells. With the strong single source held fixed, the fused arm scored worse in 17 of the 32 sealed cells before calibration and in 9 after it. Across all 72 cells of M0 the counts were 41 and 23. In one exploration task, Seoul Bike in the FULL--FULL state, fusion scored 4,323.99 before calibration against 2,098.84 for the development-selected single-source arm, and 3,089.64 against 3,116.87 after it, which reverses the comparison against that fixed reference. The reversal has two parts: calibration improved the score of the fused arm by 28.5\% and worsened that of the single-source arm by 48.5\%. Before calibration the fused intervals covered 22.7\% of the test observations and the single-source intervals 71.6\%; after it both covered more than 99.8\%, with mean widths of 3,087 and 3,117. For the fused arm the reduction in miss penalties outweighed the added width, and for the single-source arm it did not. An uncalibrated comparison cannot separate the information added by the second source from differences in interval calibration between the arms.

\subsection{Fusion gains depend on the implementation}
The verdict on fusion changed with the implementation (Figure 2). After calibration, always fusing was clearly worse than the matched single source on 5, 10, 8 and 7 of the 18 tasks under M0, M1, M2 and M3 (Table 6).

\begin{figure*}[tp]
\centering
\includegraphics[width=0.91\textwidth,height=0.70\textheight,keepaspectratio]{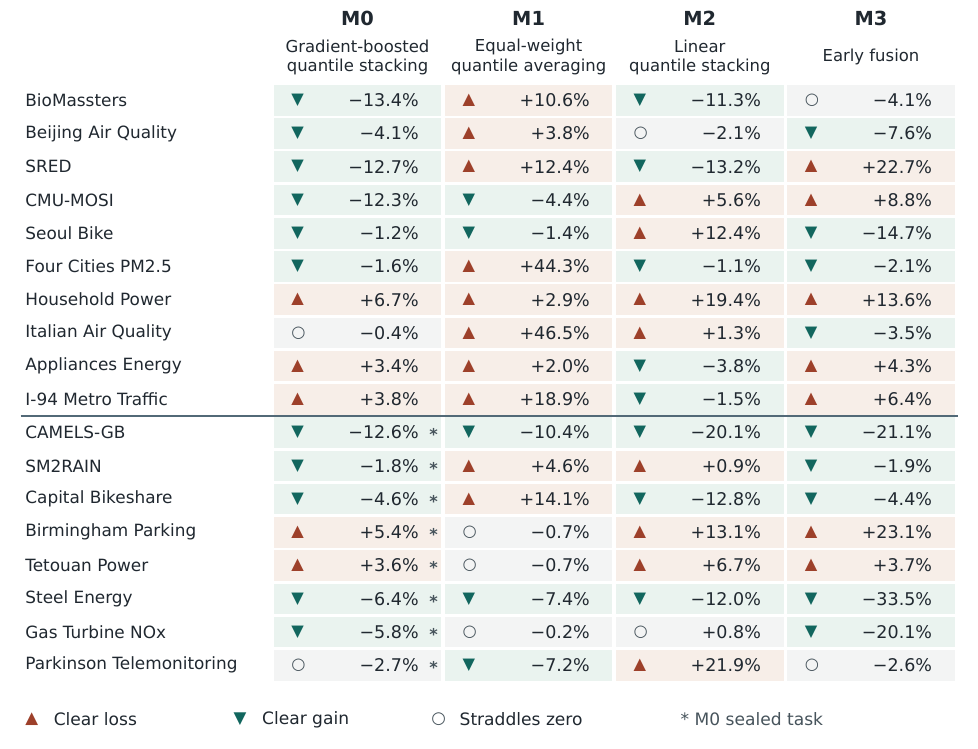}
\caption{Single-fit always-fuse verdicts under group-wise calibration. Each cell shows the relative interval-score difference, averaged equally over four retention states, against the implementation's matched strong single source. Negative values favor fusion. Upward triangles and downward triangles denote 95\% paired test-group resampling bands wholly above and below zero, respectively; open circles denote bands containing zero. Asterisks mark only the eight M0 cells in the sealed confirmation set. OpenAI Codex (client versions 0.167.0-alpha.3.2 and 0.169.0-alpha.11.1, OpenAI) assisted with writing the Python plotting code.}\label{fig:2}
\end{figure*}
\begin{table*}[tp]
\caption{Single-fit verdicts for always fusing, after calibration.}\label{tab:6}
\footnotesize
\setlength{\tabcolsep}{3pt}\renewcommand{\arraystretch}{1.13}
\begin{tabularx}{\textwidth}{@{}>{\hsize=1\hsize\raggedright\arraybackslash}X>{\hsize=1\hsize\raggedright\arraybackslash}X>{\hsize=1\hsize\raggedright\arraybackslash}X@{}}
\toprule
Implementation & All 18 tasks: clear loss / clear gain / straddles zero & Sealed 8 tasks: clear loss / clear gain / straddles zero \\
\midrule
M0 Gradient-boosted stacking & 5 / 11 / 2 & 2 / 5 / 1 \\
M1 Equal-weight averaging & 10 / 5 / 3 & 2 / 3 / 3 \\
M2 Linear stacking & 8 / 8 / 2 & 4 / 3 / 1 \\
M3 Early fusion & 7 / 9 / 2 & 2 / 5 / 1 \\
\bottomrule
\end{tabularx}
\end{table*}
The four implementations gave the same verdict on only three tasks: a clear loss on Household Power and a clear gain on CAMELS-GB and Steel Energy. On 12 of the 18 tasks, at least one implementation showed a clear loss and another a clear gain. The disagreement does not rest on equal-weight averaging, which is expected to lose when the sources differ in strength: among the three learned implementations alone, 7 tasks still showed opposite clear verdicts. Whether the second source helps is therefore not a property that can be read from the task alone. Even the identity of the stronger single source changed with the implementation in at least one state for 10 tasks.

The disagreement was not produced by the tasks that divide one table into two groups of variables. Of the 10 tasks with naturally distinct sources, 8 showed opposite clear verdicts, one showed agreement (CAMELS-GB) and one showed neither pattern (Tetouan Power). Of the 8 variable-group tasks, 4 showed opposite clear verdicts (Seoul Bike, I-94 Metro Traffic, Capital Bikeshare and Parkinson Telemonitoring), two showed agreement (Household Power and Steel Energy) and two showed neither pattern (Birmingham Parking and Gas Turbine NOx). Among the three learned implementations alone, opposite clear verdicts remained on 5 of the natural-source tasks and 2 of the variable-group tasks. Section 4.7 reports the other readouts by source type.

The task-averaged effect of always fusing ranged from a 3.15\% improvement under M0 to a 7.10\% deterioration under M1, with M2 (+0.24\%) and M3 (\ensuremath{-}1.83\%) between them. Equal-weight averaging lost most often, which is consistent with the classical result that a simple average suffers when one source is much weaker than the other.

The four values of $d$ are each measured against that implementation's own matched single source, and those references differ in strength. Table 7 puts all strategies on one reference, the development-selected single source of M0. On that reference, always fusing with equal weights was 11.8\% worse, and the selector averaged \ensuremath{-}4.7\% for M0, +0.8\% for M1, \ensuremath{-}6.4\% for M2 and \ensuremath{-}9.3\% for M3. Within M1 the comparison is matched, since neither the fused arm nor its single-source arms has a second learning stage. Its single-source reference is nonetheless weaker than a single source that has one. The 3.51\% gain of the M1 selector over its own matched reference therefore does not make it better than the common M0 reference; the two percentages have different denominators.

\begin{table*}[tp]
\caption{Task-equal mean relative interval scores against a common reference. For each task, state and fit, the denominator is the test score of the M0 single-source arm selected on development data in that run. Negative values favor the evaluated strategy.}\label{tab:7}
\footnotesize
\setlength{\tabcolsep}{3pt}\renewcommand{\arraystretch}{1.13}
\begin{tabularx}{\textwidth}{@{}>{\hsize=1\hsize\raggedright\arraybackslash}X>{\hsize=1\hsize\raggedright\arraybackslash}X>{\hsize=1\hsize\raggedright\arraybackslash}X>{\hsize=1\hsize\raggedright\arraybackslash}X>{\hsize=1\hsize\raggedright\arraybackslash}X@{}}
\toprule
Implementation & Always fuse: single fit & Always fuse: ten-refit mean & Selector: single fit & Selector: ten-refit mean \\
\midrule
M0 Gradient-boosted quantile stacking & \ensuremath{-}3.152\% & \ensuremath{-}1.035\% & \ensuremath{-}4.674\% & \ensuremath{-}4.267\% \\
M1 Equal-weight quantile averaging & +11.829\% & +11.438\% & +0.783\% & +0.552\% \\
M2 Linear quantile stacking & \ensuremath{-}2.019\% & \ensuremath{-}4.557\% & \ensuremath{-}6.434\% & \ensuremath{-}8.270\% \\
M3 Early fusion & \ensuremath{-}4.846\% & \ensuremath{-}4.891\% & \ensuremath{-}9.305\% & \ensuremath{-}9.106\% \\
\bottomrule
\end{tabularx}
\end{table*}
The same shared source models show what a conventional, unmatched comparison would have reported. M0, M1 and M2 use the same source models, so the single-source arm of M1 is a baseline without a second learning stage. For each stacking implementation, the unmatched gain over that baseline is the sum of a stage-two contrast, between the matched and the unmatched single source, and the matched gain scaled by the ratio of the two single-source scores (Table 8). Under group-wise calibration, linear stacking improved on the unmatched baseline by 5.97\% across refits and on its matched single source by 0.14\%. Nearly all of the apparent gain disappeared once the single source was given the same second learning stage and its data. For gradient-boosted stacking the two figures were 2.05\% and 1.03\%. Under pointwise calibration they were 7.73\% and 3.67\% for M0, and 9.03\% and 2.67\% for M2. In the single fit, the stage-two contrast was larger in magnitude than the matched gain on 7 to 10 of the 18 tasks.

\begin{table*}[tp]
\caption{Unmatched and matched comparisons for the two stacking implementations. The unmatched baseline is the development-selected single-source arm of M1. Negative values are improvements.}\label{tab:8}
\footnotesize
\setlength{\tabcolsep}{3pt}\renewcommand{\arraystretch}{1.13}
\begin{tabularx}{\textwidth}{@{}>{\hsize=1\hsize\raggedright\arraybackslash}X>{\hsize=1\hsize\raggedright\arraybackslash}X>{\hsize=1\hsize\raggedright\arraybackslash}X>{\hsize=1\hsize\raggedright\arraybackslash}X>{\hsize=1\hsize\raggedright\arraybackslash}X>{\hsize=1\hsize\raggedright\arraybackslash}X@{}}
\toprule
Calibration & Implementation & Unmatched gain: single fit / ten refits & Stage-two contrast: single fit / ten refits & Matched gain: single fit / ten refits & Tasks with a clear gain only against the unmatched / only against the matched baseline \\
\midrule
Group-wise & M0 & \ensuremath{-}4.73\% / \ensuremath{-}2.05\% & \ensuremath{-}1.62\% / \ensuremath{-}0.88\% & \ensuremath{-}3.15\% / \ensuremath{-}1.03\% & 4 / 4 \\
Group-wise & M2 & \ensuremath{-}4.65\% / \ensuremath{-}5.97\% & \ensuremath{-}4.50\% / \ensuremath{-}5.55\% & +0.24\% / \ensuremath{-}0.14\% & 4 / 2 \\
Pointwise & M0 & \ensuremath{-}2.13\% / \ensuremath{-}7.73\% & \ensuremath{-}4.49\% / \ensuremath{-}4.14\% & +1.58\% / \ensuremath{-}3.67\% & 5 / 0 \\
Pointwise & M2 & \ensuremath{-}9.20\% / \ensuremath{-}9.03\% & \ensuremath{-}5.90\% / \ensuremath{-}6.36\% & \ensuremath{-}3.23\% / \ensuremath{-}2.67\% & 1 / 1 \\
\bottomrule
\end{tabularx}
\end{table*}
An unmatched baseline changed verdicts in both directions. Under group-wise calibration, four tasks showed a clear gain for M0 only against the unmatched baseline and four only against the matched one; for M2 the counts were four and two. The frozen descriptive criterion for overstatement was met in all four comparisons. The stage-two contrast combines the second-stage fit with the additional stage-2-fit data and can involve a change of the selected single source, so it is not a pure effect of the learning stage.

The retention states also mattered for always fusing. When the halved source was the stronger one, always fusing compared more favorably with the matched single source than when the halved source was the weaker one. We define the stronger source as the one with the lower development score in the FULL--FULL state. This held for M1, M2 and M3 under both calibration targets; for M0 the direction differed between the two targets (Supplementary Tables S11.3 and S11.4).

\subsection{The fallback selector under four implementations and retraining}
The selector kept a low rate of harm and a positive average gain in every implementation (Table 9, Figure 3). Across 180 refit runs per implementation it scored worse than the strong single source in 9, 8, 11 and 8 runs, that is 5.0\%, 4.4\%, 6.1\% and 4.4\%. A run here is one task in one refit, averaged over its four states. The runs share their evaluation data and are not independent deployments.

\begin{figure*}[tp]
\centering
\includegraphics[width=0.91\textwidth,height=0.70\textheight,keepaspectratio]{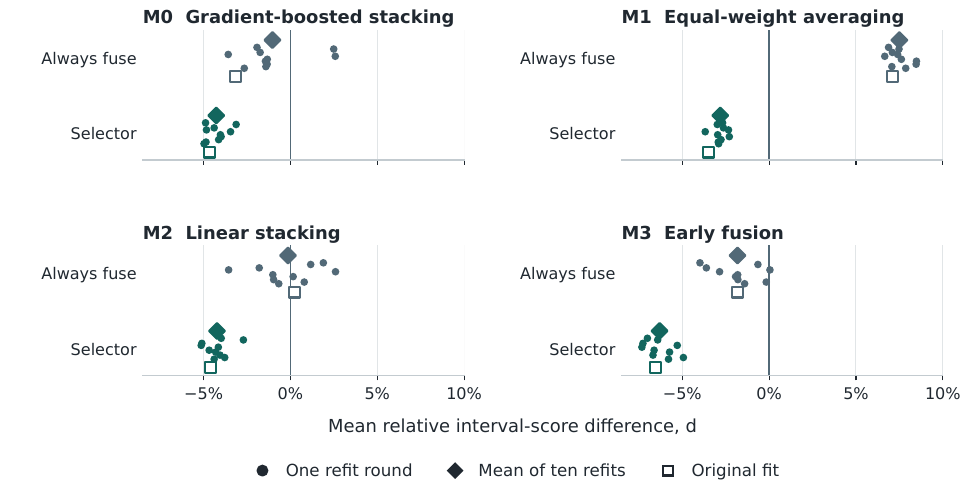}
\caption{Task-averaged relative interval-score differences across 18 tasks under group-wise calibration. Circles are the individual refit-round means, not individual task results; diamonds show the mean of ten rounds, and open squares the original fit. Vertical offsets separate the markers. Each implementation uses its own matched strong single-source reference, unlike the common M0 reference in Table 7. Negative values favor the plotted strategy. OpenAI Codex (client versions 0.167.0-alpha.3.2 and 0.169.0-alpha.11.1, OpenAI) assisted with writing the Python plotting code.}\label{fig:3}
\end{figure*}
\begin{table*}[tp]
\caption{Task-averaged $d$ for always fusing and for the selector, each against the implementation's own strong single source. Negative values are improvements. A refit round is the average over the 18 tasks for one resample.}\label{tab:9}
\footnotesize
\setlength{\tabcolsep}{3pt}\renewcommand{\arraystretch}{1.13}
\begin{tabularx}{\textwidth}{@{}>{\hsize=1\hsize\raggedright\arraybackslash}X>{\hsize=1\hsize\raggedright\arraybackslash}X>{\hsize=1\hsize\raggedright\arraybackslash}X>{\hsize=1\hsize\raggedright\arraybackslash}X@{}}
\toprule
Implementation & Always fuse: single fit / mean of 10 refits / worst refit & Selector: single fit / mean of 10 refits / worst refit & Selector runs better / equal / worse (of 180) \\
\midrule
M0 Gradient-boosted stacking & \ensuremath{-}3.15\% / \ensuremath{-}1.03\% / +2.58\% & \ensuremath{-}4.67\% / \ensuremath{-}4.27\% / \ensuremath{-}3.13\% & 131 / 40 / 9 \\
M1 Equal-weight averaging & +7.10\% / +7.50\% / +8.48\% & \ensuremath{-}3.51\% / \ensuremath{-}2.82\% / \ensuremath{-}2.29\% & 101 / 71 / 8 \\
M2 Linear stacking & +0.24\% / \ensuremath{-}0.14\% / +2.60\% & \ensuremath{-}4.58\% / \ensuremath{-}4.23\% / \ensuremath{-}2.71\% & 121 / 48 / 11 \\
M3 Early fusion & \ensuremath{-}1.83\% / \ensuremath{-}1.82\% / +0.05\% & \ensuremath{-}6.54\% / \ensuremath{-}6.31\% / \ensuremath{-}4.94\% & 116 / 56 / 8 \\
\bottomrule
\end{tabularx}
\end{table*}
In the single fit the selector chose the fused arm in 46, 29, 38 and 46 of the 72 task--state cells for M0 to M3 (63.9\%, 40.3\%, 52.8\% and 63.9\%), and across the ten refits in 56.3\%, 37.8\%, 53.1\% and 58.1\% of 720 cells. It scored strictly better than the single source in 101 to 131 of the 180 runs per implementation. Its ten-refit average improvement was 4.27\%, 2.82\%, 4.23\% and 6.31\%. In the worst of the ten rounds it still improved the average score by 3.13\%, 2.29\%, 2.71\% and 4.94\%.

Always fusing behaved differently under the same resamples. Its worst round was a deterioration in every implementation (+2.58\%, +8.48\%, +2.60\% and +0.05\%), and under M0 its ten-refit average improvement (1.03\%) was a third of its single-fit value (3.15\%). The selector turned each implementation, including the one for which always fusing lost on average, into a net improvement over that implementation's matched single source.

The selector's failures were concentrated in a few tasks. Of the 36 failing runs, 10 came from Seoul Bike and 9 from Tetouan Power. Under M0, five of the nine failures were on Seoul Bike, with losses of 0.02\% to 3.99\%; the other four were spread over three tasks and none exceeded 0.63\%. Under M1 the eight failures fell on Household Power (3), BioMassters (2) and three other tasks, with a largest loss of 1.75\%. Under M2 the eleven failures fell on Seoul Bike (4), Tetouan Power (3) and four other tasks, with a largest loss of 6.52\% on Seoul Bike. Under M3 the eight failures fell on BioMassters (3), Tetouan Power (3) and two other tasks, with a largest loss of 1.31\%.

A task-level run averages four states, and an average can hide a loss in one of them. At the level of states, the selector scored worse than the matched single source in 132 of the 2,880 comparisons across the ten refits (4.58\%). In 55 of the 720 task-level runs the task average was no worse than the single source while at least one state was (Table 10). State-level losses were small at the median and could be large in the tail: the largest was 19.96\%, under linear stacking on Parkinson Telemonitoring.

\begin{table*}[tp]
\caption{Selector outcomes at the level of task--state cells, each against the implementation's own strong single source.}\label{tab:10}
\footnotesize
\setlength{\tabcolsep}{3pt}\renewcommand{\arraystretch}{1.13}
\begin{tabularx}{\textwidth}{@{}>{\hsize=1\hsize\raggedright\arraybackslash}X>{\hsize=1\hsize\raggedright\arraybackslash}X>{\hsize=1\hsize\raggedright\arraybackslash}X>{\hsize=1\hsize\raggedright\arraybackslash}X>{\hsize=1\hsize\raggedright\arraybackslash}X@{}}
\toprule
Implementation & Single fit: better / equal / worse (of 72) & Ten refits: better / equal / worse (of 720) & Refit losses: median / 90th percentile / maximum & Task-level runs not worse on average but worse in a state (of 180) \\
\midrule
M0 Gradient-boosted stacking & 45 / 26 / 1 & 375 / 315 / 30 & 1.26\% / 5.42\% / 8.19\% & 14 \\
M1 Equal-weight averaging & 27 / 43 / 2 & 224 / 448 / 48 & 2.43\% / 8.76\% / 12.68\% & 23 \\
M2 Linear stacking & 34 / 34 / 4 & 349 / 338 / 33 & 1.45\% / 9.90\% / 19.96\% & 15 \\
M3 Early fusion & 46 / 26 / 0 & 397 / 302 / 21 & 0.57\% / 1.47\% / 5.25\% & 3 \\
\bottomrule
\end{tabularx}
\end{table*}
\subsection{Stability of a single-fit win and stability of the decision}
The extension used two prespecified stability thresholds. An implementation was labelled unstable in its single-fit verdict if fewer than 60\% of the tasks with a clear single-fit verdict kept the sign of that verdict in at least nine of ten refits, and stable in selection if the selector was worse than the strong single source in at most 10\% of refit runs. The frozen summary required at least three implementations to show both properties for a common pattern, or only the two learned stackers to show the first for a pattern specific to learned second stages. Neither condition was met, and the frozen verdict is inconsistent: all four implementations were stable in selection, and only M0 was unstable in its single-fit verdict (Table 11).

\begin{table*}[tp]
\caption{Sign stability of the single-fit verdict for always fusing under ten refits.}\label{tab:11}
\footnotesize
\setlength{\tabcolsep}{3pt}\renewcommand{\arraystretch}{1.13}
\begin{tabularx}{\textwidth}{@{}>{\hsize=1\hsize\raggedright\arraybackslash}X>{\hsize=1\hsize\raggedright\arraybackslash}X>{\hsize=1\hsize\raggedright\arraybackslash}X>{\hsize=1\hsize\raggedright\arraybackslash}X@{}}
\toprule
Implementation & Tasks with a clear single-fit verdict & Kept the sign in at least 9 of 10 refits & Share \\
\midrule
M0 Gradient-boosted stacking & 16 & 7 & 43.75\% \\
M1 Equal-weight averaging & 15 & 11 & 73.33\% \\
M2 Linear stacking & 16 & 11 & 68.75\% \\
M3 Early fusion & 16 & 12 & 75.00\% \\
\bottomrule
\end{tabularx}
\end{table*}
M2, the other learned stacker, kept the sign in at least nine refits on 68.75\% of its clear tasks. Sign changes occurred under all four implementations. Between a quarter and a third of the clear verdicts of M1, M2 and M3 also failed to hold in nine of ten refits.

Under M0 the single-fit verdict held for five tasks with a clear gain (BioMassters, Beijing, SRED, CAMELS-GB and Steel Energy) and two with a clear loss (Household Power and Birmingham Parking). It reversed often elsewhere. Seoul Bike showed a clear gain in the single fit and did not keep that sign in eight of ten refits (Supplementary Figure S1). The single-fit bands did not anticipate this. For Tetouan Power the band was 0.48 percentage points wide and the standard deviation across refits was 4.01 points; for Birmingham Parking the two figures were 2.52 and 10.94. For the tasks with stable gains the relation was reversed: the BioMassters band was 5.76 points wide and its refit standard deviation 0.97.

M0 is therefore the informative case. Its single-fit verdicts were the least stable of the four, and its selector was nonetheless worse than the single source in only 5\% of refit runs and met the sealed criterion for the selector. The stability of a single-fit win does not indicate how a selection strategy will perform, and it cannot stand in for evaluating one.

\subsection{Pre-specified gain predictors and exploratory associations}
None of the quantities available before choosing whether to fuse met the study's predictive criteria. Two hypotheses were frozen and tested in earlier phases of the project, each on a subset of the tasks. The first was that the gain from the second source is predicted jointly by the ratio of single-source scores and the correlation of their errors, as the theory of forecast combination suggests. The second was that the loss from fusion grows with the imbalance between the sources. Neither met its frozen criterion (Supplementary Table S12.1). In an exploratory analysis of all 18 tasks, none of 11 task descriptors, among them the mutual predictability of the sources, the ratio of input dimension to sample size, the gap between training and test scores and the source type, showed a rank correlation with the task-level $d$ of at least 0.5 in absolute value with the same sign in both the exploration and the sealed tasks. The ratio of the second source's input dimension to the number of source-fit rows correlated at \ensuremath{-}0.95 in the ten exploration tasks and at +0.02 in the eight sealed tasks. For the candidates tested here, a direct comparison on held-out data remained necessary.

\subsection{Sensitivity to the calibration target}
The results above use group-wise calibration, under which intervals cover almost every individual observation. As a post-opening sensitivity analysis we replaced it with signed pointwise conformal scores, the larger of $l$ \ensuremath{-} $y$ and $y$ \ensuremath{-} $u$ for each calibration observation, which target pointwise coverage and allow intervals to shrink. Predictions, roles, the scoring rule and the selection rule were unchanged, and the readouts and two descriptive criteria were frozen before computation. Median task-level pointwise coverage fell from 99.5--99.9\% to 89.1--91.0\%, and median group-wise coverage from 91.0--96.6\% to 17.3--32.8\%.

Implementation dependence persisted. The four implementations agreed on 5 of 18 tasks and gave opposite clear verdicts on 9 (Table 12). Among the three learned implementations alone, 6 tasks showed opposite clear verdicts. The calibration target itself changed verdicts: 29 of the 72 task--implementation verdicts fell in a different class under the two calibrations.

Always fusing fared better under pointwise calibration for the three learned implementations. Its ten-refit mean was \ensuremath{-}3.67\%, \ensuremath{-}2.67\% and \ensuremath{-}2.85\% for M0, M2 and M3, against \ensuremath{-}1.03\%, \ensuremath{-}0.14\% and \ensuremath{-}1.82\% under group-wise calibration, and its worst round was an improvement for M0 and M3. Equal-weight averaging moved the other way, from +7.50\% to +13.81\%. The two calibrations use different references, so this comparison is indicative only.

The selector still improved mean scores under signed pointwise calibration, but losses became more frequent. Its ten-refit mean improvement was 4.54\%, 2.16\%, 3.64\% and 5.77\% for M0 to M3, and its average over the 18 tasks was an improvement in the worst round of every implementation. It also stayed ahead of always fusing, by 0.87, 15.97, 0.97 and 2.92 percentage points in the ten-refit mean; for the two stacking implementations the margin was under one point. It scored worse than the matched single source in 39, 10, 41 and 35 of 180 task-level refit runs (21.7\%, 5.6\%, 22.8\% and 19.4\%), against 9, 8, 11 and 8 under group-wise calibration (Figure 4), and in 380 of 2,880 state-level comparisons (13.2\%). The largest state-level loss was 60.7\%. The losses were concentrated: Four Cities PM2.5 accounted for 9 of its 10 refits under both M0 and M3, and Birmingham Parking for 5 to 8 under three implementations. On the eight sealed tasks, read descriptively with M0, the selector was clearly worse than the single source on two tasks, fell back in 6 of the 13 cells where fusion scored worse, and chose fusion in 15 of the 19 where it scored better. The frozen descriptive criterion for the selector, at most 10\% of task-level runs worse in every implementation, was not met. The two calibrations led to the same choice of arm in 68.9\% of refit cells.

\begin{figure*}[tp]
\centering
\includegraphics[width=0.91\textwidth,height=0.70\textheight,keepaspectratio]{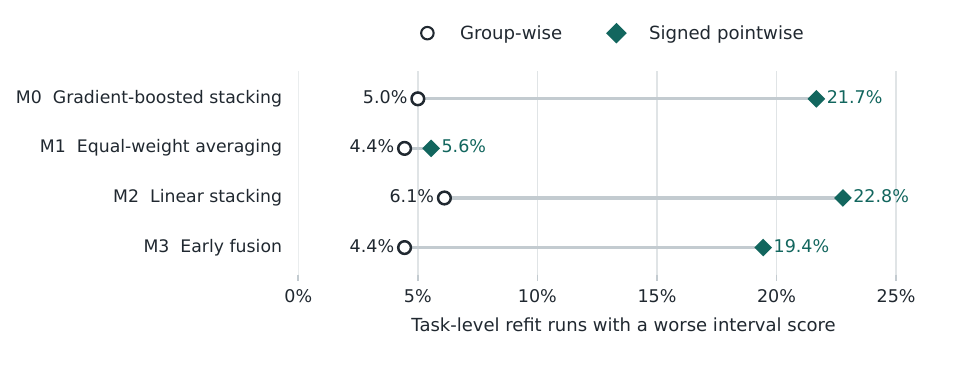}
\caption{Observed selector loss rates under group-wise and signed pointwise calibration. A loss means a higher interval score than the matched strong single source for that implementation and calibration, averaged over the four retention states. Each rate is based on 180 task-level refit runs (18 tasks in ten rounds); ties are not losses. These runs reuse evaluation data and are not independent deployments. OpenAI Codex (client versions 0.167.0-alpha.3.2 and 0.169.0-alpha.11.1, OpenAI) assisted with writing the Python plotting code.}\label{fig:4}
\end{figure*}
\begin{table*}[tp]
\caption{Main readouts under the two calibration targets. Each calibration selects its own strong single source, so the two columns of $d$ have different references.}\label{tab:12}
\footnotesize
\setlength{\tabcolsep}{3pt}\renewcommand{\arraystretch}{1.13}
\begin{tabularx}{\textwidth}{@{}>{\hsize=1\hsize\raggedright\arraybackslash}X>{\hsize=1\hsize\raggedright\arraybackslash}X>{\hsize=1\hsize\raggedright\arraybackslash}X@{}}
\toprule
Outcome & Group-wise calibration & Pointwise calibration \\
\midrule
Tasks on which the four implementations agree / give opposite clear verdicts (of 18) & 3 / 12 & 5 / 9 \\
M0 always fuse: clear loss / clear gain / straddles zero & 5 / 11 / 2 & 3 / 7 / 8 \\
M1 always fuse: clear loss / clear gain / straddles zero & 10 / 5 / 3 & 10 / 4 / 4 \\
M2 always fuse: clear loss / clear gain / straddles zero & 8 / 8 / 2 & 4 / 11 / 3 \\
M3 always fuse: clear loss / clear gain / straddles zero & 7 / 9 / 2 & 5 / 9 / 4 \\
M0 always fuse: mean refit $d$ / worst refit round & \ensuremath{-}1.03\% / +2.58\% & \ensuremath{-}3.67\% / \ensuremath{-}2.19\% \\
M1 always fuse: mean refit $d$ / worst refit round & +7.50\% / +8.48\% & +13.81\% / +15.79\% \\
M2 always fuse: mean refit $d$ / worst refit round & \ensuremath{-}0.14\% / +2.60\% & \ensuremath{-}2.67\% / +1.41\% \\
M3 always fuse: mean refit $d$ / worst refit round & \ensuremath{-}1.82\% / +0.05\% & \ensuremath{-}2.85\% / \ensuremath{-}0.69\% \\
M0 selector: mean refit $d$ / worst refit round; worse task-level runs & \ensuremath{-}4.27\% / \ensuremath{-}3.13\%; 9 of 180 & \ensuremath{-}4.54\% / \ensuremath{-}2.64\%; 39 of 180 \\
M1 selector: mean refit $d$ / worst refit round; worse task-level runs & \ensuremath{-}2.82\% / \ensuremath{-}2.29\%; 8 of 180 & \ensuremath{-}2.16\% / \ensuremath{-}1.31\%; 10 of 180 \\
M2 selector: mean refit $d$ / worst refit round; worse task-level runs & \ensuremath{-}4.23\% / \ensuremath{-}2.71\%; 11 of 180 & \ensuremath{-}3.64\% / \ensuremath{-}2.19\%; 41 of 180 \\
M3 selector: mean refit $d$ / worst refit round; worse task-level runs & \ensuremath{-}6.31\% / \ensuremath{-}4.94\%; 8 of 180 & \ensuremath{-}5.77\% / \ensuremath{-}4.68\%; 35 of 180 \\
State-level selector losses across refits (of 2,880) & 132 (4.58\%) & 380 (13.19\%) \\
\bottomrule
\end{tabularx}
\end{table*}
The pointwise analysis changes two things at once: the unit over which residuals are aggregated, and whether intervals may shrink. A third calibration separates them by using pointwise residuals with expansion only. For the three learned implementations together, the selector scored worse than the matched single source in 28 of 540 task-level refit runs under group-wise calibration, in 115 under signed pointwise calibration and in 110 under non-negative pointwise calibration. For these implementations, the bridge calibration points to observation-level residuals as the main reason for the rise in losses; prohibiting shrinkage changed the count much less. The average scores of always fusing and of the selector were also nearly unchanged between the two pointwise calibrations (Supplementary Tables S7.4 and S7.5).

\subsection{Results by source type}
Because eight tasks divide one table into two groups of variables, we report the main readouts separately for the 10 natural-source tasks and the 8 variable-group tasks (Table 13). The stratification is descriptive and was made after the sealed tasks were opened.

The main results held on the natural-source tasks. Implementations disagreed there at least as often as on the variable-group tasks. Under group-wise calibration, opposite clear verdicts occurred on 8 of the 10 natural-source tasks and 4 of the 8 variable-group tasks, and on 5 and 2 among the three learned implementations. Under pointwise calibration the counts were 5 and 4, and 3 and 3. On the natural-source tasks the selector improved the ten-refit mean score in every implementation under both calibrations, by 1.58\% to 5.05\% under group-wise calibration and by 1.34\% to 4.69\% under pointwise calibration. It scored worse than the matched single source in 3 to 7 of 100 task-level refit runs under group-wise calibration, and in 15, 4, 22 and 21 of 100 for M0 to M3 under pointwise calibration. The contrast between the two calibration targets is therefore present on naturally distinct sources alone.

The source type did change which implementation gained from always fusing. On the natural-source tasks the two stacking implementations improved on their matched single source on average under both calibrations, with ten-refit means of \ensuremath{-}3.49\% and \ensuremath{-}2.88\% under group-wise calibration and \ensuremath{-}4.03\% and \ensuremath{-}3.45\% under pointwise calibration. Early fusion did not (+1.18\% and +0.98\%). On the variable-group tasks the pattern was reversed. Early fusion improved by 5.57\% and 7.62\%, while the stacking implementations lost under group-wise calibration (+2.04\% and +3.29\%) and gained under pointwise calibration (\ensuremath{-}3.21\% and \ensuremath{-}1.70\%). Equal-weight averaging lost in both strata, most on the natural-source tasks (+11.07\% and +19.84\%). One reading is that variables from one table interact at the level of features, which early fusion can use and late fusion cannot, whereas naturally distinct sources combine well at the level of predictions. The strata contain 10 and 8 tasks and the comparison was not planned, so this reading is a conjecture.

\begin{table*}[tp]
\caption{Main readouts by source type under the two calibration targets. Verdict counts are for always fusing in the single fit. Values of $d$ are measured against each implementation's own strong single source and averaged with equal task weights within a stratum; negative values are improvements. The last column counts task-level refit runs in which the selector scored worse than the strong single source, of 100 for the natural-source tasks and 80 for the variable-group tasks.}\label{tab:13}
\scriptsize
\setlength{\tabcolsep}{3pt}\renewcommand{\arraystretch}{1.13}
\begin{tabularx}{\textwidth}{@{}>{\hsize=0.83\hsize\raggedright\arraybackslash}X>{\hsize=1.2\hsize\raggedright\arraybackslash}X>{\hsize=0.81\hsize\raggedright\arraybackslash}X>{\hsize=0.87\hsize\raggedright\arraybackslash}X>{\hsize=1.22\hsize\raggedright\arraybackslash}X>{\hsize=1.15\hsize\raggedright\arraybackslash}X>{\hsize=0.92\hsize\raggedright\arraybackslash}X@{}}
\toprule
Calibration & Implementation & Source type & Always fuse: clear loss / clear gain / straddles zero & Always fuse: single fit / ten refits & Selector: single fit / ten refits & Selector worse in task-level refit runs \\
\midrule
Group-wise & M0 & Natural & 2 / 7 / 1 & \ensuremath{-}5.19\% / \ensuremath{-}3.49\% & \ensuremath{-}6.01\% / \ensuremath{-}5.05\% & 3 of 100 \\
Group-wise & M0 & Variable-group & 3 / 4 / 1 & \ensuremath{-}0.60\% / +2.04\% & \ensuremath{-}3.01\% / \ensuremath{-}3.28\% & 6 of 80 \\
Group-wise & M1 & Natural & 7 / 2 / 1 & +10.88\% / +11.07\% & \ensuremath{-}2.09\% / \ensuremath{-}1.58\% & 3 of 100 \\
Group-wise & M1 & Variable-group & 3 / 3 / 2 & +2.38\% / +3.04\% & \ensuremath{-}5.29\% / \ensuremath{-}4.37\% & 5 of 80 \\
Group-wise & M2 & Natural & 4 / 5 / 1 & \ensuremath{-}3.70\% / \ensuremath{-}2.88\% & \ensuremath{-}5.05\% / \ensuremath{-}4.67\% & 4 of 100 \\
Group-wise & M2 & Variable-group & 4 / 3 / 1 & +5.16\% / +3.29\% & \ensuremath{-}4.00\% / \ensuremath{-}3.67\% & 7 of 80 \\
Group-wise & M3 & Natural & 4 / 5 / 1 & \ensuremath{-}0.08\% / +1.18\% & \ensuremath{-}4.10\% / \ensuremath{-}3.57\% & 7 of 100 \\
Group-wise & M3 & Variable-group & 3 / 4 / 1 & \ensuremath{-}4.01\% / \ensuremath{-}5.57\% & \ensuremath{-}9.60\% / \ensuremath{-}9.73\% & 1 of 80 \\
Pointwise & M0 & Natural & 0 / 5 / 5 & \ensuremath{-}4.62\% / \ensuremath{-}4.03\% & \ensuremath{-}4.93\% / \ensuremath{-}4.69\% & 15 of 100 \\
Pointwise & M0 & Variable-group & 3 / 2 / 3 & +9.34\% / \ensuremath{-}3.21\% & \ensuremath{-}1.65\% / \ensuremath{-}4.35\% & 24 of 80 \\
Pointwise & M1 & Natural & 7 / 2 / 1 & +20.98\% / +19.84\% & \ensuremath{-}1.56\% / \ensuremath{-}1.34\% & 4 of 100 \\
Pointwise & M1 & Variable-group & 3 / 2 / 3 & +6.73\% / +6.26\% & \ensuremath{-}3.21\% / \ensuremath{-}3.19\% & 6 of 80 \\
Pointwise & M2 & Natural & 2 / 5 / 3 & \ensuremath{-}4.22\% / \ensuremath{-}3.45\% & \ensuremath{-}5.05\% / \ensuremath{-}4.46\% & 22 of 100 \\
Pointwise & M2 & Variable-group & 2 / 6 / 0 & \ensuremath{-}2.00\% / \ensuremath{-}1.70\% & \ensuremath{-}1.16\% / \ensuremath{-}2.61\% & 19 of 80 \\
Pointwise & M3 & Natural & 3 / 4 / 3 & \ensuremath{-}1.12\% / +0.98\% & \ensuremath{-}4.13\% / \ensuremath{-}2.72\% & 21 of 100 \\
Pointwise & M3 & Variable-group & 2 / 5 / 1 & \ensuremath{-}8.78\% / \ensuremath{-}7.62\% & \ensuremath{-}10.15\% / \ensuremath{-}9.57\% & 14 of 80 \\
\bottomrule
\end{tabularx}
\end{table*}
\section{Discussion}
\subsection{Two questions with different answers}
The fused model's advantage over its matched single source varied with implementation and calibration target. For gradient-boosted stacking it also varied with the training sample. The selector improved mean scores in all four implementations under both calibrations, even though its loss rate changed. It lost in about 5\% of task-level refit runs under group-wise calibration, and in about 20\% under pointwise calibration for three implementations.

The difference between the two calibrations suggests why. Under group-wise calibration the intervals cover almost every observation, so the score of an arm is close to its width, and the width is set largely by one expansion constant that is the same on development and test data. Using the same expansion on development and test may help explain why the rankings carried over: the rank correlation between relative development and test differences on the sealed tasks was 0.95. Under pointwise calibration the score also contains miss penalties that vary from sample to sample, and the development ranking is a noisier guide.

Separating the score difference into width and miss penalties shows which component transferred more consistently. Across refits, the development-to-test rank correlation of the calibrated score difference in the original units was 0.83 to 0.88 under group-wise calibration and 0.36 to 0.82 under pointwise calibration. Under pointwise calibration the width part was ranked more consistently than the penalty part in every implementation, with correlations of 0.66 to 0.90 against \ensuremath{-}0.01 to 0.45. Of the 380 state-level losses under pointwise calibration, 229 (60.3\%) were cells in which the width difference kept its sign and a change in the penalty difference reversed the total; under group-wise calibration that share was 31.1\%, and most losses were reversals of the width difference itself. By implementation the pointwise share was 56.6\%, 70.6\%, 48.6\% and 73.9\%. The share for linear stacking fell below the frozen threshold of 50\%, so the criterion for a mechanism common to all four implementations was not met. The decomposition identifies which component accounts for a reversal in the score; it cannot identify a cause.

A fusion verdict describes a fitted pipeline on a task. Its sensitivity to refitting and calibration matters when the verdict is used to infer that the sources are complementary. On 12 of the 18 tasks the same pair of sources supported a clear gain under one implementation and a clear loss under another, and 8 of those 12 were tasks with naturally distinct sources. Agreement across implementations would give stronger grounds for attributing a gain to the sources.

The pattern fits the account of the forecast combination puzzle, in which estimated combinations lose to simpler ones because of estimation error \cite{r4,r5}. Under group-wise calibration the two learned stackers were clearly worse than their matched single sources on 5 to 8 of 18 tasks. Our design does not identify estimation error as the cause, and the sign changes under equal-weight averaging and early fusion show that learned weights are not the only source of variation.

\subsection{Consequences for reporting and using fusion}
The results suggest three practices for interval-fusion evaluations.

\begin{enumerate}
\item \textit{Compare against a matched single-source learner.} The single-source baseline should have the learning opportunity that the implementation gives the fused model: a second stage where the fused model has one, and none where it has none. Gains should also be read against their reference. Across refits, linear stacking improved on source models without a second stage by 5.97\% and on its matched single source by 0.14\% (Section 4.2).

\item \textit{Calibrate before comparing intervals.} Under gradient-boosted stacking, the number of cells in which the fused arm scored worse than a fixed single-source reference fell from 41 to 23 of 72 after calibration. The calibration target matters as well: 29 of the 72 task--implementation verdicts changed class between group-wise and pointwise calibration.

\item \textit{Report the decision as well as the win.} A fusion result should state how a rule that may return the single source performs, and how both the win and the decision behave when the model is retrained. A band computed for one fit is not evidence about the next fit.

\end{enumerate}
Include the fused arm alongside its matched single-source arms on the development set. The implementation effort is small, because the single-source arms use the same code path with one input withheld; each of them does add a model fit.

The development margin is informative about how often the selected arm also scores better on test. Across refits, the selector's choice agreed with the better test arm in 71.7\% of cells in the lowest fifth of development margins and in 99.8\% in the highest fifth under group-wise calibration, and in 57.6\% and 94.6\% under pointwise calibration. A fusion advantage that is small on development data should be treated as undecided.

\subsection{Selecting across implementations}
The selector studied here chooses within one implementation. As a descriptive extension we pooled the arms of the four implementations and chose one arm per task and state on the development data, with the strong single source of M0 as the common reference. Across refits, choosing among the 8 single-source arms gave a mean $d$ of \ensuremath{-}9.27\% under group-wise calibration and \ensuremath{-}4.86\% under pointwise calibration. Choosing among all 12 arms gave \ensuremath{-}14.23\% and \ensuremath{-}9.95\%, that is 4.96 and 5.08 percentage points more. The second source therefore added value beyond the choice of a better single-source pipeline, and by a similar amount under both calibration targets. A fused arm was chosen in about 64\% of cells, most often the early-fusion arm.

The larger pool did not improve every run. The 12-arm choice scored worse than the 8-arm choice in 11 of 180 task-level refit runs under group-wise calibration and in 34 under pointwise calibration. The single-source pipelines also differed on the common reference: the single-source arms of M2 and M3 were 4.02\% and 3.08\% better than that of M0 across refits under group-wise calibration, and that of M1 was 3.90\% worse. These comparisons were not part of any frozen criterion, and the implementations in the pool differ in training budget.

\section{Scope and limitations}
The conclusions apply to two-source regression with quantile-based prediction intervals, scored by the interval score after per-state calibration, where held-out development data are available for the choice.

\begin{itemize}
\setlength{\itemsep}{2pt}\setlength{\topsep}{3pt}
\item \textit{Confirmatory evidence covers one implementation.} The two frozen criteria were tested on eight tasks with gradient-boosted stacking under group-wise calibration. The results for the other three implementations and for all refits were obtained after the sealed tasks had been opened.

\item \textit{Refits re-use the evaluation data.} The ten refits resample the training roles only. The development, calibration and test roles are the same in every refit, so the 180 runs per implementation are not independent trials, and resampling with replacement leaves each refit with fewer distinct training groups than the original fit.

\item \textit{The implementations are not budget-matched.} Matching holds between the fused and single-source arms within an implementation. M0 uses separate source-fit and stage-2-fit roles, M1 reuses the source fits without fitting a combiner, M2 uses a capped stage-2 sample, and M3 is fitted on the uncapped union of the two fitting roles. In the SM2RAIN single fit, for example, M0 used 460,706 source-fit rows and 172,113 stage-2 rows per arm, M2 retained 11,858 stage-2 rows per arm, and M3 used 632,819 rows per arm. The values of $d$ therefore do not rank the implementations by absolute performance.

\item \textit{Eight tasks divide one table into two groups of variables.} They test the value of a second block of inputs, which is not the same as a second sensor or modality. Sections 4.2 and 4.7 report the results by source type; the main results held on the 10 natural-source tasks.

\item \textit{Temporal order is established for 14 tasks.} For four tasks the development and test data cannot be placed on a common calendar.

\item \textit{Fusion architectures.} We tested fixed quantile averaging, learned late fusion and early fusion. Whether deep multimodal architectures, attention-based fusion or per-sample gating change the disagreement between implementations remains open.

\item \textit{Thresholds are descriptive.} The 60\% and 10\% lines used to read the extension are not significance tests of differences between implementations.

\item \textit{Some tasks have few test groups.} The Parkinson Telemonitoring test role contains two patient groups, and its band is correspondingly wide.

\item \textit{The low rate of harm is specific to group-wise calibration.} Under that target the calibrated intervals covered well above 90\% of individual observations, and the scores mainly compare interval widths at a conservative level. Under pointwise calibration the selector kept its average gain but lost in about one task-level run in five for three implementations (Section 4.6). The pointwise analysis was run after the sealed tasks were opened and is descriptive.

\end{itemize}
\section{Conclusion}
Across 18 tasks and four implementations, the measured fusion gain depended on the baseline and the fitted pipeline. Under group-wise calibration, linear stacking improved the score by 5.97\% against a baseline without a second learning stage and by 0.14\% against a matched single source. With matched baselines, the implementations agreed on three tasks and gave opposite clear verdicts on twelve under group-wise calibration, seven of them among the three learned implementations alone, and agreed on five and opposed on nine under pointwise calibration. A selector that could fall back to a matched single-source model improved the task-averaged score in every implementation and under both calibrations, by 2.2\% to 6.3\% across refits, and pooling the arms of all implementations showed that the second source added about 5 percentage points beyond the choice of single-source pipeline. Under group-wise calibration the selector met the criterion frozen before eight sealed tasks were evaluated and lost in about 5\% of task-level refit runs. Under signed pointwise calibration it lost in about 20\% of runs for three implementations. Miss penalties contributed more to the score under pointwise calibration and transferred less consistently from development to test than widths did. The disagreement between implementations, the selector's average gain and the contrast in its loss rate were all present on the ten tasks with naturally distinct sources taken alone. Choose whether to fuse on held-out development data, with matched single-source arms among the candidates. Report the selector's mean gain, loss rate and size of losses on test.

\begin{samepage}
\section*{CRediT authorship contribution statement}
Liwei Deng: Conceptualization, Methodology, Software, Formal analysis, Writing - original draft. Lu Wang: Validation, Writing - review \& editing.
\par\end{samepage}

\begin{samepage}
\section*{Funding}
This work was supported by the Scientific Research Fund of Hunan Provincial Education Department (Grant No. 25C0185). The funder had no role in the study design, data collection, processing, analysis, or interpretation, manuscript preparation, or the decision to submit the article for publication.
\par\end{samepage}

\section*{Declaration of competing interests}
The authors declare that they have no known competing financial interests or personal relationships that could have appeared to influence the work reported in this paper.

\section*{Data and code availability}
The analysis code, derived result tables, protocols, and figure-generation files are not publicly released with this preprint. Raw third-party datasets and row-level prediction files are not redistributed. Dataset sources, access instructions, and preprocessing entry points are documented in Supplementary Table S2.3. CMU-MOSI data are not redistributed and should be obtained from the original provider.

\section*{Declaration of generative AI and AI-assisted technologies in the manuscript preparation process}
During preparation of this article, the authors used Claude Opus 5.5 (Anthropic) for editorial feedback and OpenAI Codex for language revision and assistance in writing analysis code. OpenAI Codex (client versions 0.167.0-alpha.3.2 and 0.169.0-alpha.11.1, OpenAI) also assisted with revising the Python plotting code. The authors reviewed and edited the AI-assisted content as needed and take full responsibility for this manuscript.


\begin{thebibliography}{99}
\bibitem{r1} W. Wang, D. Tran, M. Feiszli, What makes training multi-modal classification networks hard?, in: Proceedings of IEEE/CVF Conference on Computer Vision and Pattern Recognition, 2020, pp. 12692--12702. \url{https://doi.org/10.1109/CVPR42600.2020.01271}.
\bibitem{r2} C. Du, J. Teng, T. Li, Y. Liu, T. Yuan, Y. Wang, Y. Yuan, H. Zhao, On uni-modal feature learning in supervised multi-modal learning, in: Proceedings of International Conference on Machine Learning, PMLR 202, 2023, pp. 8632--8656.
\bibitem{r3} Q. Zhang, Y. Wei, Z. Han, H. Fu, X. Peng, Q. Hu, C. Deng, C. Xu, J. Wen, D. Hu, C. Zhang, Multimodal fusion on low-quality data: a comprehensive survey. Inf. Fusion 135 (2026) 104437. \url{https://doi.org/10.1016/j.inffus.2026.104437}.
\bibitem{r4} J. Smith, K.F. Wallis, A simple explanation of the forecast combination puzzle. Oxf. Bull. Econ. Stat. 71(3) (2009) 331--355. \url{https://doi.org/10.1111/j.1468-0084.2008.00541.x}.
\bibitem{r5} G. Claeskens, J.R. Magnus, A.L. Vasnev, W. Wang, The forecast combination puzzle: a simple theoretical explanation. Int. J. Forecast. 32(3) (2016) 754--762. \url{https://doi.org/10.1016/j.ijforecast.2015.12.005}.
\bibitem{r6} X. Bouthillier, P. Delaunay, M. Bronzi, A. Trofimov, B. Nichyporuk, J. Szeto, N. Sepah, E. Raff, K. Madan, V. Voleti, S.E. Kahou, V. Michalski, D. Serdyuk, T. Arbel, C. Pal, G. Varoquaux, P. Vincent, Accounting for variance in machine learning benchmarks. Proc. Mach. Learn. Syst. 3 (2021) 747--769.
\bibitem{r7} M. Hibon, T. Evgeniou, To combine or not to combine: selecting among forecasts and their combinations. Int. J. Forecast. 21(1) (2005) 15--24. \url{https://doi.org/10.1016/j.ijforecast.2004.05.002}.
\bibitem{r8} N. Bosch, O. Shchur, N. Erickson, M. Bohlke-Schneider, A.C. Turkmen, Multi-layer stack ensembles for time series forecasting, in: Proceedings of International Conference on Automated Machine Learning, PMLR 293, 2025, pp. 15/1--30.
\bibitem{r9} Z. Xue, R. Marculescu, Dynamic multimodal fusion, in: Proceedings of IEEE/CVF Conference on Computer Vision and Pattern Recognition Workshops, 2023, pp. 2575--2584.
\bibitem{r10} C. Qu, Z. Lu, Y. Lai, T.Z. Li, J. Zhu, J. Guo, J. Xiong, Y. Zhu, Y. Yang, A.J. Luna, K.L. Sandler, B.A. Landman, Y. Huo, AdaFuse: adaptive multimodal fusion for lung cancer risk prediction via reinforcement learning, in: Proceedings of Medical Imaging with Deep Learning, PMLR 315, 2026, pp. 2582--2601.
\bibitem{r11} F. Krones, T. Papastylianou, G. Qian, G. Parsons, B. Shukla, B.W. Papież, A. Mahdi, Advancing ICU decision making through multimodal data fusion in predicting cardiovascular disorders. Inf. Fusion 136 (2026) 104573. \url{https://doi.org/10.1016/j.inffus.2026.104573}.
\bibitem{r12} I. Azizi, Conformal calibration for multi-modal regression with missing modalities, in: Proceedings of Conformal and Probabilistic Prediction with Applications (COPA), PMLR 329, 2026, pp. 343--377.
\bibitem{r13} A. Bose, J. Ethier, P. Guinand, Conformal prediction for multimodal regression. arXiv:2410.19653 (2024). \url{https://doi.org/10.48550/arXiv.2410.19653}.
\bibitem{r14} D. Gkoumas, Q. Li, C. Lioma, Y. Yu, D. Song, What makes the difference? An empirical comparison of fusion strategies for multimodal language analysis. Inf. Fusion 66 (2021) 184--197. \url{https://doi.org/10.1016/j.inffus.2020.09.005}.
\bibitem{r15} P.P. Liang, Y. Lyu, X. Fan, Z. Wu, Y. Cheng, J. Wu, L. Chen, P. Wu, M.A. Lee, Y. Zhu, R. Salakhutdinov, L.-P. Morency, MultiBench: multiscale benchmarks for multimodal representation learning, in: Proceedings of the Neural Information Processing Systems Track on Datasets and Benchmarks, vol. 1, 2021.
\bibitem{r16} I. Burenko, D. Vetrov, Beyond modality fusion: deep ensembles for multimodal classification. arXiv:2607.05019 (2026). \url{https://doi.org/10.48550/arXiv.2607.05019}.
\bibitem{r17} J.M. Bates, C.W.J. Granger, The combination of forecasts. Oper. Res. Q. 20(4) (1969) 451--468. \url{https://doi.org/10.1057/jors.1969.103}.
\bibitem{r18} S.M. Blanc, T. Setzer, When to choose the simple average in forecast combination. J. Bus. Res. 69(10) (2016) 3951--3962. \url{https://doi.org/10.1016/j.jbusres.2016.05.013}.
\bibitem{r19} B. Schulz, L. Köhler, S. Lerch, Aggregating distribution forecasts from deep ensembles. Mach. Learn. 115 (2026) 120. \url{https://doi.org/10.1007/s10994-025-06946-3}.
\bibitem{r20} Y. Yang, A.K. Kuchibhotla, Selection and aggregation of conformal prediction sets. J. Am. Stat. Assoc. 120(549) (2025) 435--447. \url{https://doi.org/10.1080/01621459.2024.2344700}.
\bibitem{r21} R. Liang, W. Zhu, R.F. Barber, Conformal prediction after data-dependent model selection. J. Am. Stat. Assoc. (2026), advance online publication. \url{https://doi.org/10.1080/01621459.2026.2663588}.
\bibitem{r22} I. Azizi, I. Rudnytskyi, Improving real estate rental estimations with visual data. Big Data Cogn. Comput. 6(3) (2022) 96. \url{https://doi.org/10.3390/bdcc6030096}.
\bibitem{r23} T. Gneiting, A.E. Raftery, Strictly proper scoring rules, prediction, and estimation. J. Am. Stat. Assoc. 102(477) (2007) 359--378. \url{https://doi.org/10.1198/016214506000001437}.
\end{thebibliography}
\end{document}